\documentclass[10pt, a4paper]{article}
\usepackage{amssymb,amsfonts,bm}
\usepackage{makecell}

\usepackage{latexsym}
\usepackage{graphicx} 
\usepackage{microtype}
\usepackage{multirow}
\usepackage{multicol}
\usepackage{booktabs}

\usepackage{lrec-coling2024} 

\title{\textsc{NSina}: A {N}ews Corpus for {Sin}hal{a}}

\name{Hansi Hettiarachchi\textsuperscript{1}, Damith Premasiri\textsuperscript{2}, Lasitha Uyangodage\textsuperscript{3}, \\ \large \textbf{Tharindu Ranasinghe\textsuperscript{4}}} 

\address{\textsuperscript{1}Birmingham City University, UK, \textsuperscript{2}Lancaster University, UK, \\ \textsuperscript{3}University of Münster, Germany, \textsuperscript{4}Aston University, UK \\
         \textsuperscript{1}\texttt{hansi.hettiarachchi@bcu.ac.uk}, \textsuperscript{2}\texttt{d.dolamullage@lancaster.ac.uk}, \\ \textsuperscript{3}\texttt{luyangod@uni-muenster.de}, \textsuperscript{4}\texttt{t.ranasinghe@aston.ac.uk}}

\abstract{
The introduction of large language models (LLMs) has advanced natural language processing (NLP), but their effectiveness is largely dependent on pre-training resources. This is especially evident in low-resource languages, such as Sinhala, which face two primary challenges: the lack of substantial training data and limited benchmarking datasets. In response, this study introduces \textsc{NSina}, a comprehensive news corpus of over 500,000 articles from popular Sinhala news websites, along with three NLP tasks: news media identification, news category prediction, and news headline generation. The release of \textsc{NSina} aims to provide a solution to challenges in adapting LLMs to Sinhala, offering valuable resources and benchmarks for improving NLP in the Sinhala language. \textsc{NSina} is the largest news corpus for Sinhala, available up to date. 
 \\ \newline  }

\begin{document}

\maketitleabstract

\section{Introduction}
\vspace{-4mm}
The recent emergence of large language models (LLMs) has ushered in significant advancements in the field of natural language processing (NLP) \cite{devlin-etal-2019-bert}. These LLMs have produced state-of-the-art results in many NLP benchmarks, outperforming previous machine learning models such as LSTMs \cite{lin2022survey}. Beyond academic research, these recent LLMs, such as GPT \cite{brown2020language}, have also driven the development of widely popular products, including chatbots, machine translation and writing assistants, among other applications, making them highly popular among the general public.

While LLMs have achieved considerable success and garnered popularity, their effectiveness heavily relies on access to resources for weight pre-training. Consequently, these LLMs excel in high-resource languages but encounter challenges when applied to low-resource languages \cite{wang-etal-2020-extending}. Two primary factors contribute to the complexities of deploying LLMs in low-resource linguistic contexts: (1) Limited availability of open-access corpora for pre-training the models in low-resource languages. and (2) The absence of appropriate benchmarking datasets in these languages to assess the performance of the models. In this research, we address these challenges for Sinhala by releasing \textsc{NSina}: A large {N}ews Corpus for {Sin}hala accompanied by a range of benchmarking tasks. 

Sinhala is an Indo-Aryan language spoken by over 17 million people in Sri Lanka. Sinhala is one of the two official languages in Sri Lanka. Predominantly, the Sinhala-speaking community comprises the Sinhalese people, the largest ethnic group on the island. Despite its sizable community, Sinhala remains relatively under-resourced compared to other languages spoken in the region \cite{de2019survey}. The complexities regarding LLMs and low-resource languages that we discussed previously appear in Sinhala too. While there exist several multilingual language models, such as XLM-RoBERTa \cite{conneau-etal-2020-unsupervised}, which offer support for Sinhala, the multilingual corpora used to train these models contain a relatively limited proportion of Sinhala compared to other languages \cite{wang-etal-2020-extending}. For example, the OSCAR 23.01 multilingual corpus \citelanguageresource{abadji2022towards}, which has been used to train multilingual language models, only contains 2.6GB Sinhala text which does not even contribute to 1\% of the total dataset. Furthermore, since these data were automatically extracted from Common Crawl dumps, they contain noise resulting from boilerplate content extracted from headers, footers and sidebars of web crawls \citelanguageresource{abadji2022towards}.


There is also a pre-trained BERT model trained specifically for Sinhala \cite{sinhala_bert}. However, it's worth noting that this model has been trained on a relatively small Sinhala corpus, leading to its inability to consistently outperform multilingual LLMs in various Sinhala NLP tasks, as evidenced in both \citet{sinhala_bert} and \citetlanguageresource{ranasinghe2022sold}. These constraints primarily stem from the limited availability of sizable Sinhala corpora for model training. Furthermore, as we mentioned before, there is a huge limitation of available benchmarking datasets for Sinhala. This is also evident in \citet{sinhala_bert}, where they evaluated the pre-trained BERT model in only three text classification tasks.

In this paper, we lay the foundation to address these challenges. Firstly, we compiled a large news corpus with more than 500,000 news articles from ten popular news websites in Sri Lanka. Secondly, we compose three NLP tasks from the news corpus, including two text classification tasks: (1) News media identification (2) News category prediction and one text generation task: (3) News headline generation. We release separate train and test sets for each task, which are sampled from \textsc{NSina}. These datasets can be used as benchmarks to evaluate the LLMs in Sinhala. While there are several news corpora for Sinhala \citelanguageresource{upeksha2015implementing}, \textsc{NSina} is the most updated and the largest Sinhala news corpus available.

Our \textbf{main contributions} are;

\begin{enumerate}
    \item We introduce \textsc{NSina}, a large {N}ews Corpus for {Sin}hal{a}, which comprises 506,932 news articles, and we describe the steps taken to compile it. 

    \item We introduce three NLP tasks associated with \textsc{NSina}. (i) News media identification, (ii) News category prediction, and (iii) News headline generation. We release train and test sets for each task, which are subsampled from \textsc{NSina}.

    \item We evaluate several machine learning models on each task and compare the performance. 

    \item We release \textsc{NSina}, as an open-access publicly available dataset alongside the trained machine-learning models for each task\footnote{The dataset is available at \url{https://github.com/Sinhala-NLP/NSINA}}. 
\end{enumerate}

\section{Dataset Construction}
We first present the data collection methodology we used followed by a detailed statistical analysis of \textsc{NSina}.

\subsection{Data Collection}
First, we identified ten news sources that are popular in Sri Lanka. These sources encompass a diverse range, including ``Adaderana," ``ITN News," ``Lankatruth," ``Divaina," ``Hiru News," ``Lankadeepa," ``Vikalpa," ``Dinamina," and ``Siyatha." Importantly, we took deliberate steps to ensure a balanced representation within this selection, encompassing both pro-government and anti-government news outlets.

We used a Python script to scrape each news media website. For each news article, we extracted the source, timestamp, headline, news content, URL and category.

\subsection{\textsc{NSina}}
After data collection, we performed data cleaning. There were a considerable amount of news articles that had less than ten Sinhala words. These articles were identified and filtered. The final dataset consisted of 506,932 news articles. The source news media of these news articles is shown in Table \ref{tab:nsina}. As can be seen, ``Lankadeepa" and ``Hiru News" contribute most to the corpus, having more than 100,000 articles each. 

\begin{table}[!ht]
\centering
\scalebox{1.00}{
\begin{tabular}{|l|r|}
    \hline
    \textbf{Source} & \textbf{Amount}  \\
    \hline
    Adaderana & 83918 \\
    ITN News & 30777 \\
    Lankatruth & 48180  \\
    Divaina  & 26043  \\
    Hiru News & 130729 \\
    Sinhala News LK  & 20371\\
    Lankadeepa & 141663 \\
    Vikalpa & 14309 \\
    Dinamina & 7642 \\
    Siyatha &  3300 \\

    \hline
    Total & 506932 \\
    \hline
\end{tabular}
}
\caption{Number of news articles from each source in \textsc{NSina}.}
\label{tab:nsina}
\end{table}

\begin{figure}
\centering
\includegraphics[scale=0.35]{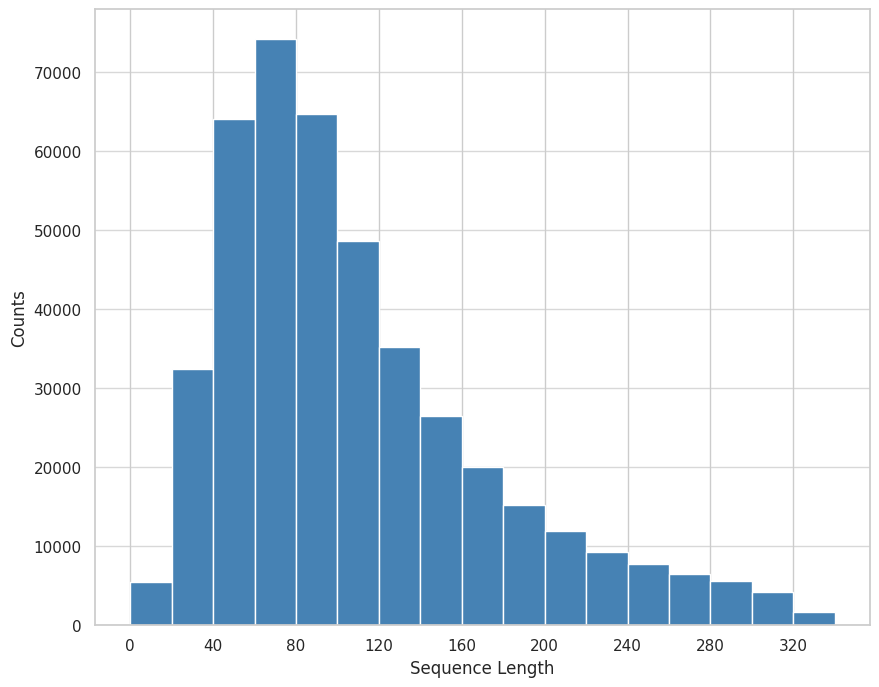}
\caption{Token frequency distribution of news content in \textsc{NSina}}
\label{fig:frequency_distribution}
\end{figure}

\begin{figure}
\centering
\includegraphics[scale=0.4]{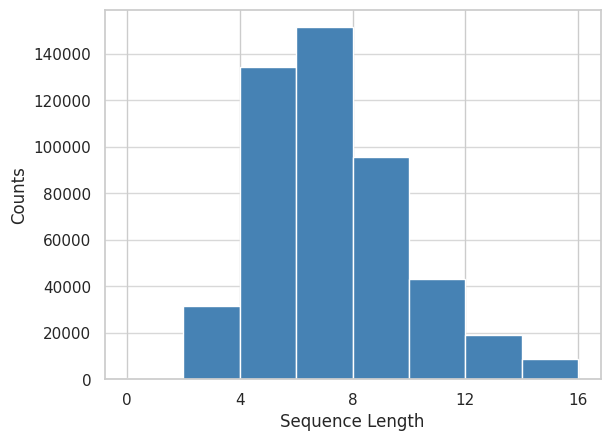}
\caption{Token frequency distribution of news headline in \textsc{NSina}}
\label{fig:frequency_distribution_title}
\end{figure}

Figure \ref{fig:frequency_distribution} shows the token frequency distribution of the news content after removing the outliers (very long documents). It is clear that most news articles contain around 60-120 tokens. Furthermore, in Figure \ref{fig:frequency_distribution_title}, we plot the token frequency of headlines. The figure shows that most of the titles are short, having between 6-12 tokens.  The corpus has more than 1 million tokens and more than 100,000 unique tokens. 

Previous Sinhala news corpus; SinMin \cite{upeksha2015implementing} was only 1.01 GB in size. Compared to that, \textsc{NSINa} is 1.87 GB and provides a larger and updated news corpus. Furthermore, the Sinhala portion of the OSCAR 23.01 corpus \cite{abadji2022towards} only has 301,066 documents and compared to that \textsc{NSINa} has more than 500,000 properly cleaned Sinhala documents.

\section{Tasks}

We compiled the following three tasks from \textsc{NSina}. We believe that creating benchmarks would help to evaluate the LLMs in Sinhala NLP tasks. 

\begin{enumerate}
\vspace{-2mm}
    \item News media identification
    \vspace{-2mm}
    \item News category prediction
    \vspace{-2mm}
    \item News headline generation
    \vspace{-2mm}
\end{enumerate}

Each of the tasks had different train/ test sets and machine learning models. The following sub-sections will describe tasks, machine learning models trained for the task and the results. 

\subsection{News Media Identification} 
The first task we compiled from \textsc{NSina} is identifying news source given the news content. This task serves a dual purpose: firstly, it aids in recognising the distinctive style of each news outlet in their news presentation. Moreover, this task can be further explored to unveil potential political biases within Sri Lankan news media. This is a text classification task, where the input to the ML models would be the news content, and the expected output of the model is the news media name. 

\subsubsection{Data}
As all the news instances in \textsc{NSina} contained its news source, constructing the train/ test set for this task was straightforward. However, as can be seen in Table \ref{tab:nsina}, since some news media sources have more instances, we undersampled them. We only used 10,000 instances from each news source. For the two sources that had less than 10,000 instances ("Dinamina" and "Siyatha") we used the original number of instances they contained. We divided this dataset into a training and test set following a 0.8 split\footnote{The sampled dataset is available at \url{https://github.com/Sinhala-NLP/Sinhala-News-Media-Identification}}.  

\subsubsection{Models}
For this task, we experimented with several text classification models based on transformers. From an input sentence, transformers compute a feature vector $\bm h\in\mathbb{R}^{d}$, upon which we build a classifier for the task. For this task, we implemented a softmax layer, i.e., the predicted probabilities are $\bm y^{(B)}=softmax(W\bm h + b)$, where $W\in\mathbb{R}^{k\times d}$ is the softmax weight matrix, and $k$ is the number of labels.

For these, we employed a batch-size of 16, Adam optimiser with learning rate $2\mathrm{e}{-5}$, and a linear learning rate warm-up over 10\% of the training data. During the training process, the parameters of the transformer model, as well as the parameters of the subsequent layers, were updated. The models were evaluated at least three times per epoch while training using an evaluation set that had one-fifth of the rows in training data. We performed early stopping if the evaluation loss did not improve over three evaluation steps. All the models were trained for three epochs. 

\subsubsection{Results}
We evaluate the models using the average weighted F1 score and macro F1 score. The results are shown in Table \ref{tab:task1}.

\begin{table}[!ht]
\centering
\scalebox{1.00}{
\begin{tabular}{|l|c|c|}
    \hline
    \textbf{Models} & \textbf{Weighted F1} & \textbf{Macro F1}  \\
    \hline
     \texttt{XLM-R Large} & 0.8917  & 0.8763 \\
         \texttt{XLM-R Base} & 0.8961 & 0.8854 \\
    \texttt{SinBERT Small} & 0.8966 & 0.8883 \\
    \texttt{SinBERT Large} & \textbf{0.8967} & \textbf{0.8895} \\

    \hline
\end{tabular}
}
\caption{News media classification results. The results are ordered with the ascending order for \textbf{Macro F1} score.}
\label{tab:task1}
\end{table}

As can be seen in the results, the performance of all the transformer models on the task was excellent. All the models achieved more than 0.88 Macro F1 score on the task. These outcomes imply that each news source exhibits a distinct stylistic approach in presenting their news content.

Notably, the \texttt{SinBERT Large} model outperformed the \texttt{XLM-R Large} model. However, the \texttt{XLM-R Large} model produces very close results to the \texttt{SinBERT Large} model.

\subsection{News Category Prediction}
The second task we compiled from \textsc{NSina} was news category prediction. This is an old task in NLP, but it has many potential applications \cite{bracewell2009category}. Given the news content, the ML models should predict a pre-defined category for the news. 

\subsubsection{Data}
Preparing the train/test sets for this task was more challenging than the previous task as the categories are not commonly defined for all news media. First, for this task, we dropped all the news articles without a category as some news sources prefer not to categorise them. Next, we identified the top 100 news categories from the available news instances. We grouped them into four main categories: local news, international news, sports news, and business news. To avoid bias, we undersampled the dataset. We only used 10,000 instances from each category, and for the ``Business" category, we used the original number of instances which was 8,777 articles. We divided this dataset into a training and test set following a 0.8 split\footnote{The sampled dataset is available at \url{https://github.com/Sinhala-NLP/Sinhala-News-Category-Prediction/}}.

\subsubsection{Models}
We employed the same ML models experimented with for the previous task, as this is also a text classification task. The hyperparameter configurations and training strategy also remained the same. 

\subsubsection{Results}
Similar to the previous task, we used Weighted F1 and Macro F1 scores to evaluate the models. The results are shown in Table \ref{tab:task2}.

\begin{table}[!ht]
\centering
\scalebox{1.00}{
\begin{tabular}{|l|c|c|}
    \hline
    \textbf{Models} & \textbf{Weighted F1} & \textbf{Macro F1}  \\
    \hline
    \texttt{SinBERT Small} & 0.9310 & 0.9306 \\
    \texttt{SinBERT Large} & 0.9370 & 0.9367 \\
    \texttt{XLM-R Large} & 0.9385  & 0.9381 \\
    \texttt{XLM-R Base} & \textbf{0.9387} & \textbf{0.9382} \\

    \hline
\end{tabular}
}
\caption{News category prediction results. The results are ordered with the ascending order for \textbf{Macro F1} score. The best results are in bold.}
\label{tab:task2}
\end{table}

Similar to the previous task, transformer models provided excellent results in this task as well. This can be a particularly easy task as the news content within each category we analysed is distinct and varied. Notably, \texttt{XLM-R} models could outperform \texttt{SinBERT} model in this task. 

\subsection{News Headline Generation}
Lastly, we created a natural language generation (NLG) task using the \textsc{NSina} corpus, where the ML model's objective is to generate news headlines based on the provided news content. As all the news content in \textsc{NSina} has headlines, constructing this NLG task was straightforward. In an era of increasing interest in language generation models such as GPT, this benchmark will be invaluable for evaluating their performance in the context of Sinhala language generation.

\subsubsection{Data}
We used the same instances from \textsc{NSina} as all the news articles had headlines. We divided this dataset into a training and test set following a 0.8 split\footnote{The sampled dataset is available at \url{https://github.com/Sinhala-NLP/Sinhala-Headline-Generation}}.

\subsubsection{Models}
In this task, we exploit two types of ML models based on transformers.

\paragraph{General Transformers} -  We created a Seq2Seq model from general transformers by adding a transformer decoder, which takes the encoder's output and generates the target sequences. We experimented with several general-purpose transformer models that support Sinhala, including XLM-Roberta \cite{conneau-etal-2020-unsupervised}, and SinBERT \cite{sinhala_bert}.

\paragraph{Text Generation Transformers} - We also experimented with several text generation transformers as they have provided excellent results in text generation tasks. Specifically, we explored mBART \cite{lewis-etal-2020-bart} and several mT5 \cite{xue-etal-2021-mt5} variants. Both mBART and mT5 support Sinhala. 

For both types of transformer models, we employed a batch size of 16, Adam optimiser with learning rate $1\mathrm{e}{-4}$, and a linear learning rate warm-up over 10\% of the training data. During the training process, the parameters of the transformer model were updated. The models were trained only using the training data and evaluated while training using an evaluation set that had one-fifth of the rows in training data. We performed early stopping if the evaluation loss did not improve over three evaluation steps. All the models were trained for three epochs.

\subsubsection{Results}

We evaluated the results of the models using two popular NLG evaluation metrics: BLEU \cite{papineni2002bleu} and Translation Edit Rate (TER). While there are advanced NLG metrics such as BLEURT \cite{sellam-etal-2020-bleurt} and BERTScore \cite{zhang2019bertscore}, they do not currently support Sinhala. 

\begin{table}[!ht]
\centering
\scalebox{1.0}{
\begin{tabular}{|l|c|c|}
    \hline
    \textbf{Models} & \textbf{BLEU} & \textbf{TER}  \\
    \hline
    \texttt{SinBERT} & 0.08 & 0.80 \\
    \texttt{XLM-R Base} & 0.11 & 0.76\\
    \texttt{XLM-R Large} & 0.14  & 0.75 \\\hline
    \texttt{mBART} & 0.15 & 0.74 \\
    \texttt{mT5 Base} & 0.16 & 0.73\\
    \texttt{mT5 Large} & 0.17  & 0.72 \\
    \hline
\end{tabular}
}
\caption{News category prediction results. The results are ordered in ascending order for BLEU.}
\label{tab:task3}
\end{table}

The results in Table \ref{tab:task3} show that all the transformer models do not perform well in this natural language processing task. The mT5-large model produced the best result with a BLEU score of 0.17. However, a BLEU score in the range of 0.1 and 0.2 suggests that there is a poor overlap between the generated headline and the actual headline.

The poor performance on NLG can be attributed to two main reasons. (1) There is no model trained specifically for Sinhala language generation and (2) The BLEU and TER scores cannot properly evaluate Sinhala text generation and there is a need for advanced NLG metrics for Sinhala. 

\section{Conclusion}
This research presented \textsc{NSina}, a large news corpus for Sinhala that can be used to train LLMs. \textsc{NSina} is larger and more recent than previous news corpus released for Sinhala, such as SinMin \cite{upeksha2015implementing}. We also release three benchmark datasets sampled from \textsc{NSina} that can be used to evaluate LLMs. We evaluated several transformer models on each task. The results show that multilingual transformer models such as XLM-R provide very close results or sometimes even outperform language-specific models such as SinBERT \cite{sinhala_bert}, suggesting more research should be done to train Sinhala-specific transformer models. Furthermore, all the experimented models perform poorly on the proposed NLG task, suggesting that more language generation models should be explored for Sinhala. 

In future work, we would like to utilise \textsc{NSina} with other available Sinhala resources to create robust transformer models. Furthermore, a GLUE \cite{wang-etal-2018-glue} like benchmark will be created for Sinhala, including the tasks proposed in this paper, which will serve as a pivotal platform for evaluating the proficiency of LLMs in processing Sinhala.

\section*{Acknowledgments} 

The computational experiments in this paper were conducted on the Aston EPS Machine Learning Server, funded by the EPSRC Core Equipment Fund, Grant EP/V036106/1.

\section*{Ethics Statement} 

\textsc{NSina} was collected from publicly available websites, and none of the records were edited in the process. Similar to the previous research that compiled and released news corpora \citelanguageresource{nagoudi-etal-2020-machine, kakwani-etal-2020-indicnlpsuite}, for every instance in \textsc{NSina}, we released the URL for the original news article. Furthermore, we released \textsc{NSina} with the Creative Commons Attribution-Non Commercial-ShareAlike 4.0 International Public License, which prevents users from altering the instances in the dataset. While \textsc{NSina} is publicly available in HuggingFace, we only released it as a gated dataset where the users need to accept the license and request access. All the datasets released for the subtasks in this paper also follow a similar process. 

\nocite{*}
\section{Bibliographical References}\label{sec:reference}

\bibliographystyle{lrec-coling2024-natbib}
\bibliography{lrec-coling2024-example}

\section{Language Resource References}
\label{lr:ref}
\bibliographystylelanguageresource{lrec-coling2024-natbib}
\bibliographylanguageresource{languageresource}

\end{document}